\documentclass[letterpaper, 10 pt, conference]{ieeeconf}  

\IEEEoverridecommandlockouts                              

\title{\LARGE \bf
TransGraspNet: Physically and Geometrically Consistent Manipulation of Transparent Labware
}

\author{Hailing Hu$^{1}$, Mingyi Zhu$^{2}$, Yiquan An$^{1}$, Yifei Tian$^{1}$, Tianyou Zuo$^{1}$, Jian S. Dai$^{3}$, and Lifeng Zhou$^{1,*}$
\thanks{*Corresponding author: Lifeng Zhou (email: lifengzhou@pku.edu.cn; ORCID: 0000-0001-5479-3681)}%
\thanks{$^{1}$Hailing Hu, Yiquan An, Yifei Tian, Tianyou Zuo, and Lifeng Zhou are with Peking University, Beijing, China (Lifeng Zhou is also with the School of Advanced Manufacturing and Robotics, Peking University).}%
\thanks{$^{2}$Mingyi Zhu is with Shanghai Jiao Tong University, Shanghai, China.}%
\thanks{$^{3}$Jian S. Dai is with Southern University of Science and Technology, Shenzhen, China.}%
}

\usepackage[backend=biber, style=ieee]{biblatex}
\usepackage{amsmath}
\usepackage{amsfonts}  
\usepackage{amssymb}   
\usepackage{amsmath,amsfonts}
\usepackage{algorithmic}
\usepackage{algorithm}
\usepackage{array}
\usepackage[caption=false,font=normalsize,labelfont=sf,textfont=sf]{subfig}
\usepackage{textcomp}
\usepackage{stfloats}
\usepackage{url}
\usepackage{verbatim}
\usepackage{graphicx}

\usepackage[backend=biber, style=ieee]{biblatex}
\AtBeginBibliography{%
  \setlength{\itemsep}{0pt}%
  \setlength{\parsep}{0pt}%
  \setlength{\parskip}{0pt}%
}         

\begin{document}

\maketitle
\thispagestyle{empty}
\pagestyle{empty}

\begin{abstract}

Manipulating transparent laboratory glassware that contains liquid is inherently safety-critical: even small geometric errors can cause unstable grasps and hazardous spillage. Although recent progress has been made in transparent object perception and robotic grasping, most existing systems optimize detection, depth reconstruction, and grasp planning independently, which leads to cross-stage inconsistency—imperfect boundaries induce depth bleeding, distorted surfaces corrupt normal estimation, and task-agnostic grasp scoring yields tilted or off-center grasps that fail under dynamic motion. In this paper, we propose TransGraspNet, a geometry–physics consistent framework that explicitly enforces consistency from perception to execution through three coupled principles: boundary consistency to produce structurally reliable object contours as downstream priors, surface consistency to preserve geometric fidelity and surface normal accuracy during depth reconstruction, and physics consistency to refine grasp selection with centroid alignment and wrench-space stability for upright and dynamically robust manipulation. We evaluate TransGraspNet on public benchmarks, a dedicated transparent glassware dataset, and a real robotic platform. The results show improved boundary quality and surface normal fidelity, and demonstrate strong task-level performance in cluttered transparent scenes. Most importantly, the proposed system achieves reliable real-world operation, including high grasp success rates in clutter and zero spillage during high-speed liquid transport, highlighting its practicality for safety-critical laboratory automation.

\end{abstract}

\section{Introduction}

Autonomous laboratory systems, often referred to as “robot scientists,” aim to perform complex experimental procedures with minimal human intervention. In such environments, robotic manipulators must reliably handle transparent laboratory glassware such as beakers, cylinders, and test tubes under cluttered and reflective conditions. Unlike structured industrial parts, transparent objects exhibit severe optical ambiguity: refractions, specular highlights, and background distortions dominate RGB observations, while commodity depth sensors frequently suffer from missing returns and boundary corruption. When these objects contain liquid, the manipulation problem becomes inherently safety-critical—minor geometric errors can induce tilted grasps, unstable transport, or even hazardous spillage. Therefore, robust and physically stable transparent object manipulation is a fundamental requirement for laboratory automation.

Despite recent advances in transparent object perception and robotic grasping, most existing systems adopt a cascaded pipeline consisting of detection, depth completion, and grasp planning, where each component is optimized independently. While such modular designs perform reasonably well in isolated benchmarks, they often fail in transparent manipulation scenarios. The underlying reason is error propagation caused by cross-stage geometric inconsistency. Inaccurate object boundaries lead to depth bleeding during reconstruction; distorted depth surfaces degrade surface normal estimation; and grasp planners that rely solely on local geometric feasibility produce tilted or off-center grasps. In cluttered laboratory environments, these inconsistencies accumulate across stages and ultimately manifest as unstable physical interaction.

We argue that the core challenge in transparent glassware manipulation is not merely the need for stronger models within individual modules, but the lack of geometry–physics consistency across perception and action stages. Transparent objects amplify the consequences of structural misalignment between modules: a boundary error in perception can fundamentally alter the reconstructed surface geometry, which in turn compromises force-closure analysis and dynamic stability during execution. Addressing this problem therefore requires a principled framework that enforces structural and physical consistency throughout the entire pipeline.

To this end, we propose TransGraspNet, a geometry–physics consistent framework for transparent object manipulation. At the perception stage, we introduce boundary-consistent reasoning to generate structurally reliable object contours that serve as explicit geometric priors for downstream reconstruction. During depth reconstruction, we enforce surface consistency through boundary-guided gating and geometry-preserving constraints to prevent depth bleeding and maintain surface normal fidelity. Finally, at the manipulation stage, we incorporate physics-consistent grasp refinement by integrating centroid alignment and wrench-space stability analysis, transforming geometrically feasible grasps into dynamically safe and upright executions suitable for liquid-containing glassware.

The contributions of this work are fourfold. First, we establish a unified geometry–physics consistent framework that systematically addresses cross-stage inconsistency in transparent object manipulation. Second, we design a boundary-enforced reconstruction strategy that preserves geometric structure and surface normal fidelity under severe optical ambiguity. Third, we develop a physics-aware grasp refinement mechanism that explicitly incorporates task-level stability constraints for safe execution. Finally, we validate the proposed system on public benchmarks and a real robotic platform, including dynamic liquid transport experiments, demonstrating robust performance in cluttered transparent laboratory environments.

\begin{figure}[H]
\centering
\includegraphics[width=2in]{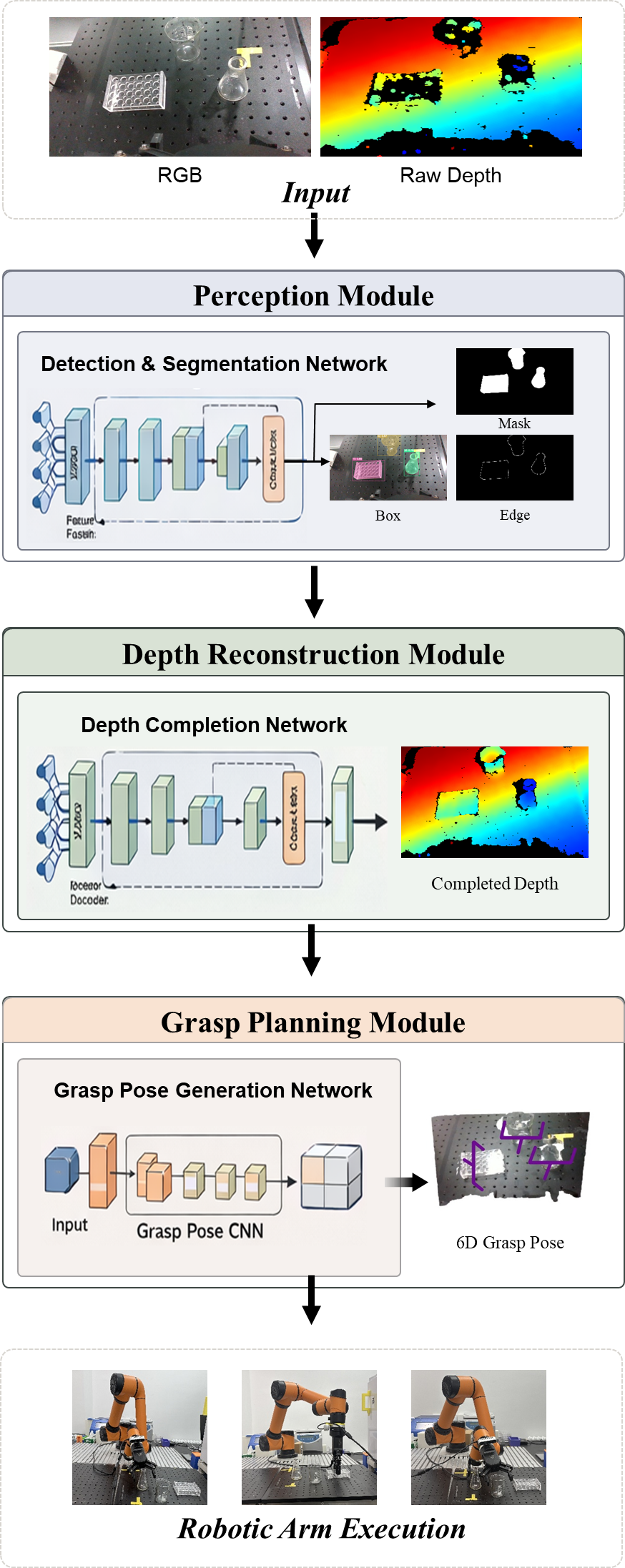} 
\caption{Overall framework. TransGraspNet consist of perception, depth reconstruction, and grasp planning modules to ensure cross-stage consistency for transparent labware manipulation.}
\label{11x}
\end{figure}


\section{Related Work}

\subsection{Transparent Object Detection and Segmentation}
Transparent objects lack inherent texture, making their appearance heavily dependent on background and lighting, which defeats traditional color- or texture-based methods. Early work exploited physics-driven cues: specular highlights \cite{osadchy2003using}, edge distortion \cite{mchenry2005finding}, light-field consistency \cite{xu2015transcut}, and polarization \cite{kalra2020deep}. While effective in controlled settings, these approaches require specialized hardware or strict lighting, limiting generalization in open laboratories.

With deep learning, data-driven methods became mainstream. Lai et al. \cite{lai2015transparent} first applied R-CNN with color and highlight features. Sajjan et al. \cite{sajjan2020clear} used Mask R‑CNN on synthetic data for instance segmentation. Xie et al. released the Trans10K dataset and the dual‑branch TransLab model \cite{xie2020segmenting}, later improving with Trans2Seg, a Transformer‑based approach treating segmentation as dictionary lookup \cite{xie2021trans2seg}.

Despite progress, challenges remain for robotic manipulation. First, state‑of‑the‑art methods focus on semantic segmentation (e.g., “glass” category), whereas grasping requires high‑precision instance segmentation to distinguish individual objects. Second, applying generic instance models like Mask R‑CNN to transparent objects suffers from “perspective effects”—background textures leak into masks, causing blurred edges or holes. To address this, we refine Mask R‑CNN with CBAM attention for noise suppression and an explicit Edge Branch to enforce geometric contours, achieving fine‑grained perception at low cost.

\subsection{Transparent Object Depth Completion and Restoration}
Transparent objects cause severe depth corruption in consumer RGB-D cameras due to infrared transmission and specular reflection, creating an ill-posed reconstruction problem that requires additional visual cues.

Early depth completion research targeted autonomous driving, using sparse convolutions \cite{uhrig2017sparsity} and spatial propagation networks like CSPN \cite{cheng2018depth} and NLSPN \cite{park2020non}. While effective for LiDAR data, these methods fail on transparent objects where valid depth seeds are rarely acquired from glass surfaces.

Addressing transparent objects specifically, ClearGrasp \cite{sajjan2020clear} predicted normals, boundaries, and masks for global optimization, but its multi-stage pipeline is too slow for real-time grasping. Later methods improved shape coherence via implicit neural representations \cite{zhu2021rgb} and occlusion awareness \cite{fang2022transcg}, though inference costs remain high in cluttered scenes.

The current efficient paradigm is RGB-guided depth completion, where high-resolution RGB guides sparse depth filling \cite{ma2018sparse, tang2020learning, qiu2019deeplidar}. However, these methods face severe modality misalignment in laboratory scenarios. Multi-modal fusion networks \cite{hu2019revisiting, jaritz2018sparse} assume RGB and depth edges are spatially consistent—an assumption broken for transparent objects. Background textures seen through glass create texture-structure ambiguity, causing networks to produce depth bleeding or over-smoothing \cite{tang2020learning, zhu2021rgb}.

To address this, we propose TDCNet with a gating mechanism \cite{hu2018squeeze, li2020multi} that adaptively learns RGB feature confidence and suppresses guidance in texture-rich but geometrically flat regions, achieving high-fidelity reconstruction.

\subsection{Robotic Grasp Detection and Planning}
Robotic grasping generates feasible grasp configurations from perceptual input, broadly categorized into planar and 6-DoF grasping.

\subsubsection{Planar Grasping}
Early planar grasping used analytic methods for force closure constraints \cite{bicchi2000robotic}. Deep learning advanced this field: Lenz et al. \cite{lenz2015deep} used cascaded networks for grasp rectangles; Redmon et al. \cite{redmon2015real} introduced YOLO-based single-stage detection; Morrison et al. \cite{morrison2018closing} proposed GG-CNN for real-time grasp quality maps. While computationally efficient, planar grasping restricts robots to top-down operations, limiting applicability in laboratory scenarios requiring side-grasping of beakers.

\subsubsection{6-DoF Grasping}
Six-DoF grasping enables complete 3D pose inference. The sample-and-evaluate paradigm, exemplified by GPD \cite{ten2017grasp} and PointNetGPD \cite{liang2019pointnetgpd}, samples candidates and evaluates contact quality. Dex-Net \cite{mahler2017dex} leveraged physics simulation for robustness prediction. Recent advances include GraspNet-1Billion with VAE-based generation and geometric scoring \cite{fang2020graspnet}, Contact-GraspNet for dense prediction via implicit maps \cite{sundermeyer2021contact}, and Transformer \cite{liu2022transgrasp} or diffusion-based \cite{zuo2024graspdiff} models for handling transparent or reflective objects.

\subsubsection{Task-Agnostic vs. Task-Constrained Grasping}
Generic 6-DoF models excel in geometric stability but remain task-agnostic, with scoring functions optimizing only friction and contact depth \cite{fang2020graspnet, sundermeyer2021contact}. In robot scientist applications, grasping must also satisfy process constraints: pipetting requires upright orientation, and grasp points must clear liquid centroids to prevent tipping. Directly applying generic GraspNet often yields high-confidence but tilted poses. This paper addresses this gap by embedding laboratory-specific physics safety constraints into the scoring module, selecting physically safe grasps from geometrically feasible candidates.

\section{Method}

We propose \textbf{TransGraspNet}, a geometry--physics consistent framework for transparent glassware manipulation. The system enforces structural coherence from 2D boundary perception to 3D surface reconstruction and finally to physics-aware grasp execution. Given RGB-D input $(I, D)$, the pipeline outputs a safety-compliant 6-DoF grasp pose.

\subsection{Edge-Guided Boundary Consistency}

Transparent objects exhibit weak internal texture but relatively clear refractive contours. To enhance boundary reliability, we build upon a ResNet-101 + FPN backbone and introduce an Enhanced CBAM (E-CBAM) attention module followed by a dual-stream boundary head.

The attention refinement is defined as:
\begin{equation}
F' = M_s(M_c(F) \otimes F) \otimes (M_c(F) \otimes F),
\end{equation}
where $M_c$ and $M_s$ denote channel and spatial attention, respectively, and $\otimes$ is element-wise multiplication.

A lightweight Edge Branch predicts contour probability $P_{edge}$, which is fused with mask features:
\begin{equation}
F_{fused} = \text{Concat}(F_{mask}, F_{edge}).
\end{equation}

Edge supervision is generated via morphological erosion:
\begin{equation}
G_{edge} = G_{mask} - \text{Erosion}(G_{mask}).
\end{equation}

The overall loss is:
\begin{equation}
L_{total} = L_{cls} + L_{box} + L_{mask} + \lambda L_{edge},
\end{equation}

with Dice-style edge loss:
\begin{equation}
L_{edge} = 1 - 
\frac{2 \sum_i p_i g_i + \epsilon}
{\sum_i p_i + \sum_i g_i + \epsilon}.
\end{equation}

The predicted mask $M_{obj}$ and edge map $E_{mask}$ are explicitly forwarded as geometric priors for depth reconstruction.

\begin{figure}[H]
\centering
\includegraphics[width=3.5in]{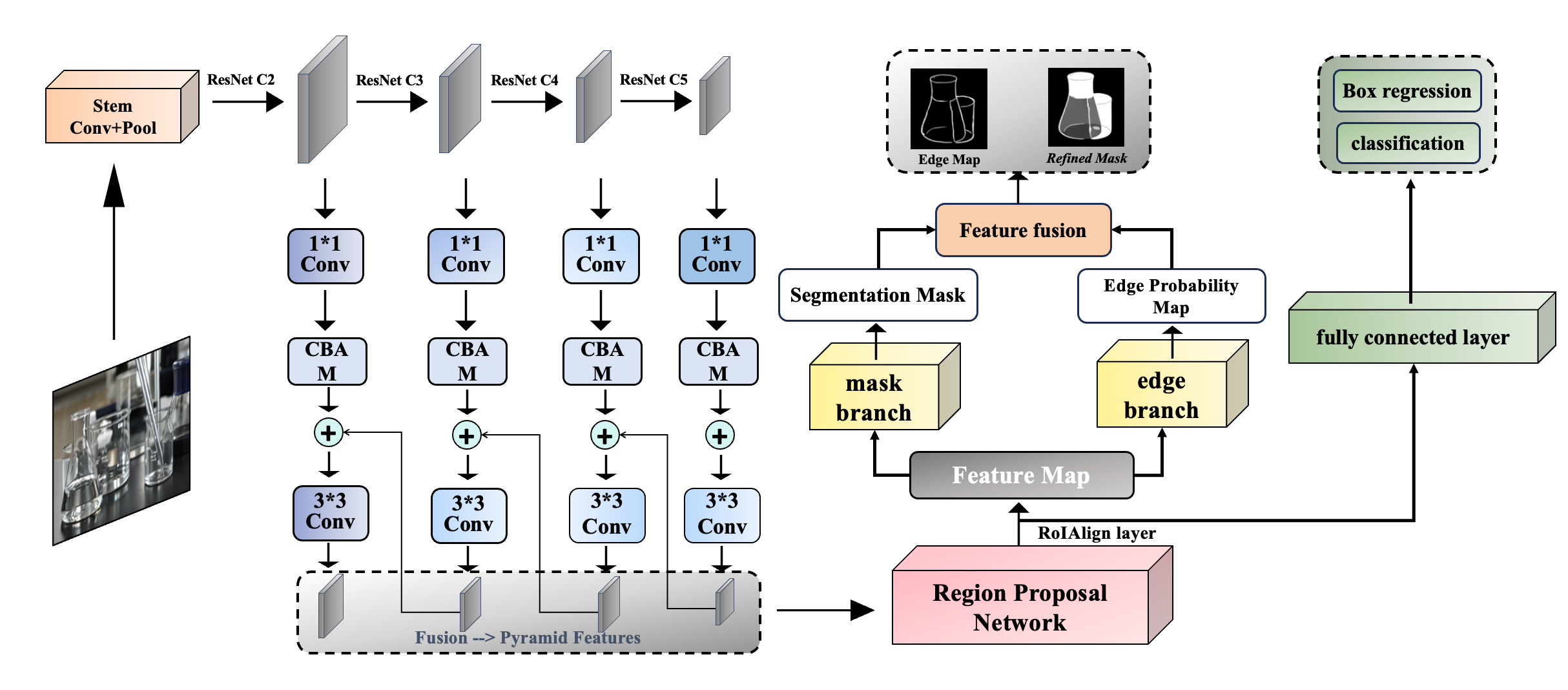} 
\caption{Architecture of TransGraspNet-Det. The network employs E-CBAM for edge-aware feature enhancement and a dual-stream head for joint boundary-guided segmentation and object detection.}
\label{xxxxx}
\end{figure}

\subsection{Surface Consistency via Geometry-Aware Depth Completion}

Depth sensing on transparent objects suffers from missing returns and cross-boundary diffusion. We adopt a TDCNet backbone and introduce an Edge-Guided Attention Gate (EGAG) to regulate multi-modal fusion.

At scale $l$, fusion is defined as:
\begin{equation}
\alpha_{gate} = \sigma(W_g \cdot \text{Concat}(F_{rgb}, F_{depth}, E_{mask})),
\end{equation}

\begin{equation}
\tilde{F}^{(l)} = \phi(F_{depth}^{(l)}, \alpha_{gate} \odot F_{rgb}^{(l)}).
\end{equation}

This mechanism suppresses cross-boundary depth propagation while preserving interior smoothness.

To preserve object geometry, we introduce a Masked Geometric Retention (MGR) loss:
\begin{equation}
L_{smooth} =
\sum_{p \in \Omega}
(\alpha_{high} M_{obj}(p) + \alpha_{decay} M_{bg}(p))
|\nabla D(p)|.
\end{equation}

Surface normal accuracy is evaluated as:
\begin{equation}
\theta_{err} =
\frac{1}{|\Omega|}
\sum_{i \in \Omega}
\arccos
\left(
\frac{n_i^{pred} \cdot n_i^{gt}}
{\|n_i^{pred}\| \|n_i^{gt}\|}
\right).
\end{equation}

The refined depth $D_{refined}$ is back-projected to generate point cloud $P$ and surface normals $N$.

\begin{figure}[H]
\centering
\includegraphics[width=4in]{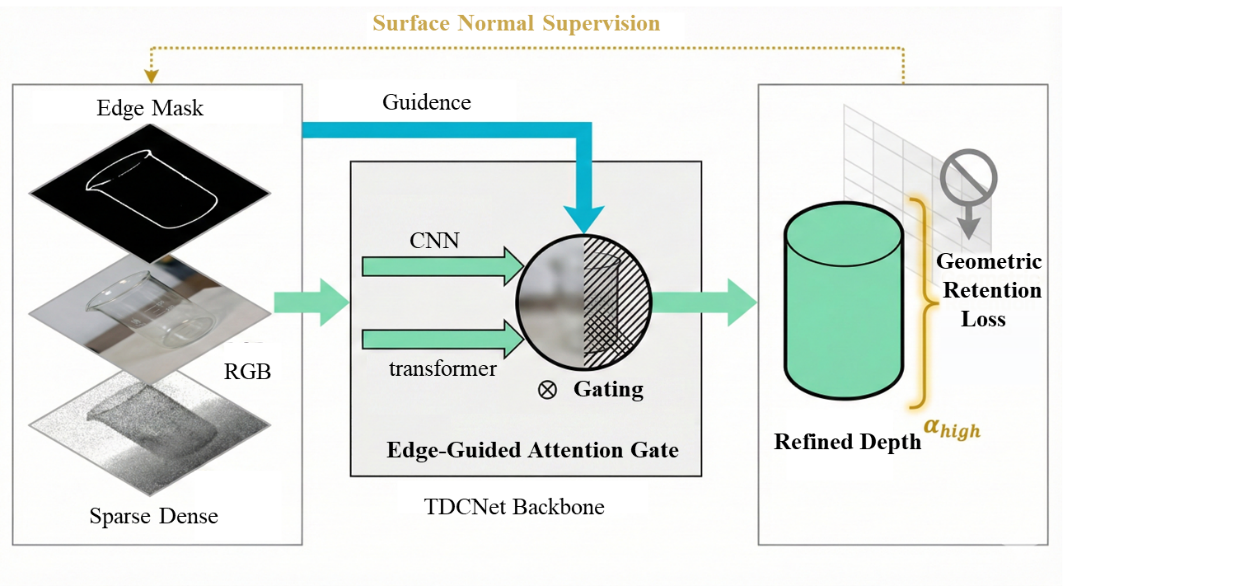} 
\caption{Architecture of TransGraspNet-Depth. The model integrates an Edge-Guided Attention Gate (EGAG) and Masked Geometric Retention (MGR) loss to suppress depth bleeding and preserve surface normal fidelity.}
\label{xxxxxx}
\end{figure}

\subsection{Geometry--Physics Aware Grasp Refinement}

Given grasp candidates $\{g_i\}$ generated by a pre-trained 6-DoF model, we refine their scores by integrating geometric alignment and physical stability.

Using PCA, we compute object centroid $c_{obj}$ and principal axis $v_{main}$. For grasp $g_i$ with center $c_g$ and approach vector $v_{app}$, we define:

\textbf{Radial alignment:}
\begin{equation}
S_{radial} =
\exp\left(
-\frac{d(c_g, v_{main})^2}{2\sigma_r^2}
\right),
\end{equation}

\textbf{Angular matching:}
\begin{equation}
S_{angle} =
\exp\left(
-\frac{(\theta_{app}-\theta_{ideal})^2}{2\sigma_a^2}
\right),
\end{equation}

\textbf{Centroid alignment:}
\begin{equation}
S_{centroid} =
\exp\left(
-\frac{\Delta_{axial}^2 + \Delta_{lateral}^2}{2\sigma_c^2}
\right).
\end{equation}

At the contact level, we evaluate:

\textbf{Antipodal condition:}
\begin{equation}
S_{anti} = \mathbb{I}(\pi - \theta_n < 2\beta),
\end{equation}

\textbf{Wrench-space robustness:}
\begin{equation}
S_Q = \text{Radius}(\text{ConvexHull}(\text{Wrenches})).
\end{equation}

The final score is:
\begin{equation}
\begin{split}
& w_0 S_{\text{raw}} + w_1 S_{\text{radial}} + w_2 S_{\text{angle}} \\
&\quad + w_3 S_{\text{centroid}} + w_4 S_{\text{anti}} + w_5 S_Q = S_{\text{final}}.
\end{split}
\end{equation}

\begin{figure}[H]
\centering
\includegraphics[width=3.3in]{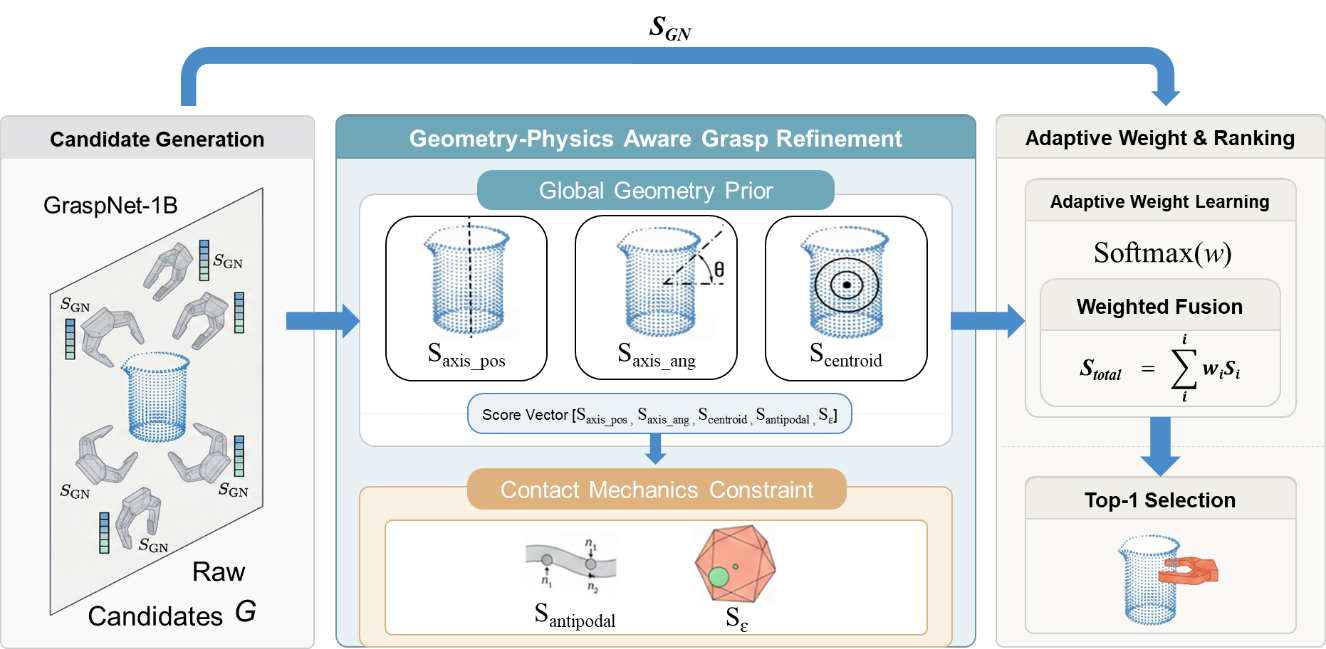} 
\caption{Geometry-physics aware grasp refinement. Raw candidates are re-scored by fusing global geometric priors (axis/centroid alignment) with contact mechanics (antipodal and wrench-space Q1 metrics) via adaptive weight learning.}
\label{xxxxxxxx}
\end{figure}

\subsection{System Execution}

The selected grasp $g^*$ is transformed to the robot base frame and executed through a standard approach--grasp--lift--place pipeline. By enforcing boundary, surface, and physics consistency across stages, the system produces upright and dynamically stable grasps suitable for liquid-containing glassware.

\section{Experiments and Results}
We validate TransGraspNet on transparent perception, depth restoration, and grasp planning via public benchmark evaluation, domain fine-tuning on RobotSci-Glass, and real-robot grasping. We further conduct ablations to isolate the contribution of key modules.

\subsection{Dataset Preparation}
To address scarce transparent-object data and unavailable depth ground truth in chemical labs, we adopt \textbf{Public Benchmark Pre-training + Domain-Specific Fine-tuning}.

\subsubsection{Public Benchmarks}
\begin{itemize}
    \item \textbf{Trans10K:} 10k+ real images for pre-training TransGraspNet-Det to learn generic transparent appearance cues.
    \item \textbf{ClearGrasp:} synthetic+real RGB-D for pre-training TransGraspNet-Depth to learn depth recovery priors from noisy measurements.
\end{itemize}

\subsubsection{RobotSci-Glass Dataset}
We construct RobotSci-Glass for ``Robot Scientist'' operations, covering 20 categories (15 transparent glassware). We use \textbf{Paired Acquisition \& Sparse Fine-tuning} with two subsets:
\begin{enumerate}
    \item \textbf{Perception Subset:} 5,000+ RGB-D images under varying backgrounds/lighting, annotated with 2D instance masks to train and test segmentation robustness.
    \item \textbf{Depth Golden Subset:} 200 representative scenes for accurate depth supervision. We apply \textbf{Opaque Coating} to blacken transparent vessels, enabling a RealSense camera to capture near noise-free depth as Ground Truth. This subset is used exclusively for fine-tuning and quantitative evaluation of TransGraspNet-Depth.
\end{enumerate}

\subsection{Implementation Details}
All models are implemented in PyTorch on Ubuntu 20.04 with an NVIDIA RTX 4090 GPU.

\subsubsection{Perception Network}
We use ResNet-101 (ImageNet initialized). Training follows a two-stage protocol: pre-train on Trans10K for 20 epochs (lr=0.002) to learn generic transparent features, then fine-tune on RobotSci-Glass Perception Subset for 10 epochs (lr=0.0002) to improve laboratory vessel boundaries.

\subsubsection{Depth Completion Network}
Based on TDCNet with ClearGrasp pre-trained weights, we fine-tune only on the Depth Golden Subset. To avoid overfitting, we apply a \textbf{Partial Freezing Strategy}: freeze the first three encoder layers while updating the decoder and the proposed EGAG. We use AdamW with batch size 8.

\subsubsection{Grasp Scoring Module}
We generate initial 6D candidates using the official GraspNet-1Billion model, then apply our geometric-physics constrained refinement as post-processing. We calibrate the score weights offline using 200 typical grasps labeled success/failure and fit a linear regression to maximize correlation with grasp success (no backprop training required).

\subsection{Evaluation Metrics}
We evaluate each module with task-aligned metrics.

\subsubsection{Perception Metrics}
We report $AP_{box}$ and $AP_{mask}$. Since transparent boundaries are easily blurred, we additionally use \textbf{Boundary IoU}, computed only within a narrow contour band to better reflect fine-edge quality.

\subsubsection{Depth and Geometry Metrics}
We report RMSE, MAE, and REL for depth accuracy. To assess whether the recovered surface is physically plausible (e.g., cylindrical curvature), we use \textbf{Mean Surface Normal Error ($\theta_{err}$)}:
\begin{equation}
    \theta_{err} = \frac{1}{|\Omega|} \sum_{i \in \Omega} \arccos \left( \frac{\mathbf{n}^{pred}_i \cdot \mathbf{n}^{gt}_i}{\| \mathbf{n}^{pred}_i \| \| \mathbf{n}^{gt}_i \|} \right)
\end{equation}
where $\Omega$ denotes valid pixels. Lower $\theta_{err}$ indicates smoother and more geometry-consistent reconstructions, which is critical for force-closure analysis.

\subsubsection{Grasp Success Rate}
A successful grasp completes a full loop: lift, transport, and stable placement (no slip/collision). We compute:
\begin{equation}
    \text{Success Rate} = \frac{N_{success}}{N_{total}} \times 100\%
\end{equation}
where $N_{total}$ is the total trials and $N_{success}$ is the number of successes.

\begin{figure}[H]
\centering
\includegraphics[width=3.3in]{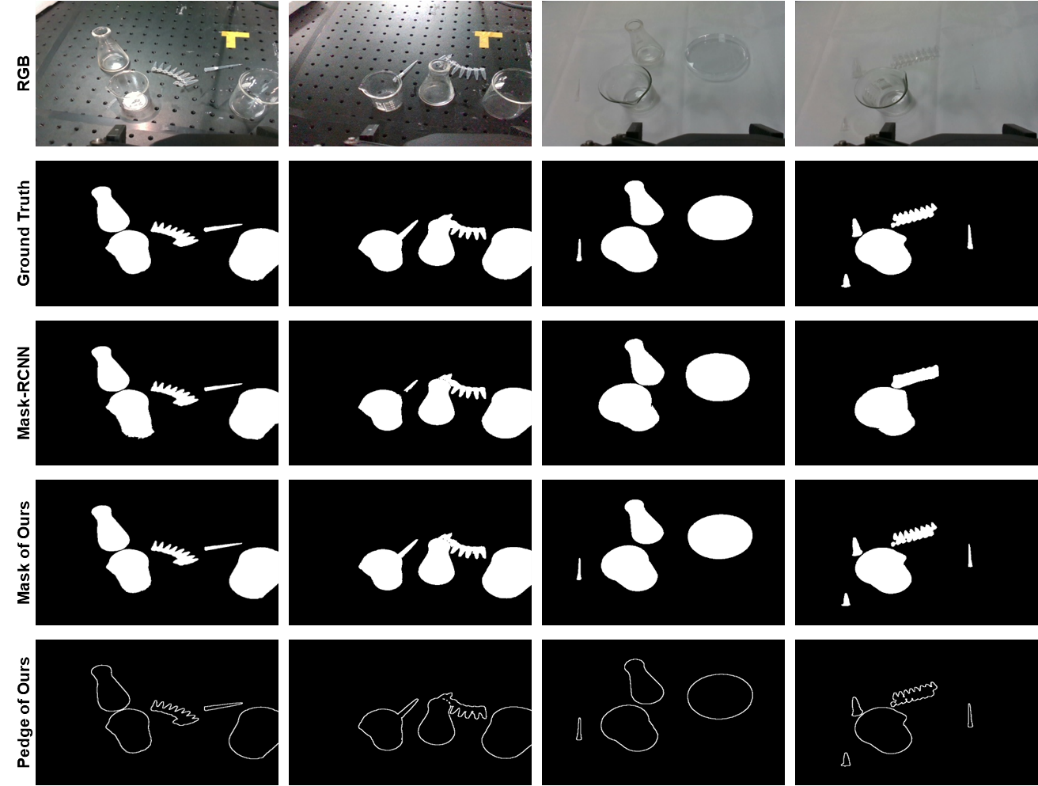} 
\caption{Qualitative comparison of edge-guided mechanism. Compared to Mask R-CNN, our method with E-CBAM and edge supervision effectively suppresses reflection noise and preserves complete contours.}
\label{4}
\end{figure}

\subsection{Component Effectiveness Analysis}
We validate the necessity of key modules via ablations and qualitative visualizations.

\subsubsection{Impact of Edge-Guided Perception}
We conduct orthogonal ablations on RobotSci-Glass to evaluate the \textbf{E-CBAM} and the \textbf{Edge Branch}.

\begin{table}[!t]
\caption{Ablation Study on Instance Segmentation Performance}
\label{tab:ablation_seg}
\centering
\resizebox{\columnwidth}{!}{
\begin{tabular}{c l c c c c}
\hline
ID & Method & E-CBAM & Edge Branch & $AP_{mask}$ (\%) & Boundary F (\%) \\
\hline
1 & Baseline (Mask R-CNN) & $\times$ & $\times$ & 72.5 & 48.2 \\
2 & + Attention Only & \checkmark & $\times$ & 74.1 (+1.6) & 51.5 (+3.3) \\
3 & + Edge Branch Only & $\times$ & \checkmark & \textbf{75.9 (+3.4)} & \textbf{62.8 (+14.6)} \\
4 & Ours (Full Method) & \checkmark & \checkmark & \textbf{78.5 (+6.0)} & \textbf{65.3 (+17.1)} \\
\hline
\end{tabular}%
} 
\end{table}

As shown in Table~\ref{tab:ablation_seg}, E-CBAM improves overall segmentation mainly by suppressing background-induced false positives, while the Edge Branch is the dominant factor for contour accuracy (+14.6\% Boundary F). Combining both yields the best performance, because attention reduces reflective high-frequency clutter and allows edge supervision to focus on true object boundaries.

Figure~\ref{4} confirms that the baseline errors largely come from reflection noise causing mask bleeding, while our edge-guided constraints produce more reliable masks for downstream depth completion.

\subsubsection{Geometry-Aware Depth Completion}
We compare our depth module (EGAG + MGR loss) against the original TDCNet.

\begin{table}[!t]
\caption{Quantitative Comparison of Depth Completion on RobotSci-Glass}
\label{tab:depth_ablation}
\centering
\begin{tabular}{l c c c}
\hline
Method & RMSE (mm) $\downarrow$ & REL $\downarrow$ & Normal Error ($^\circ$) $\downarrow$ \\
\hline
Baseline (TDCNet) & 25.4 & 0.09 & 15.2 \\
Ours & \textbf{18.1 (-7.3)} & \textbf{0.06} & \textbf{8.4 (-6.8)} \\
\hline
\end{tabular}
\end{table}

As shown in Table~\ref{tab:depth_ablation}, our method reduces RMSE by 7.3\,mm and, more importantly, halves the normal error (15.2$^\circ$ $\rightarrow$ 8.4$^\circ$), indicating that we recover physically consistent curvature rather than only smoothing pixels.

\begin{figure}[H]
\centering
\includegraphics[width=3.3in]{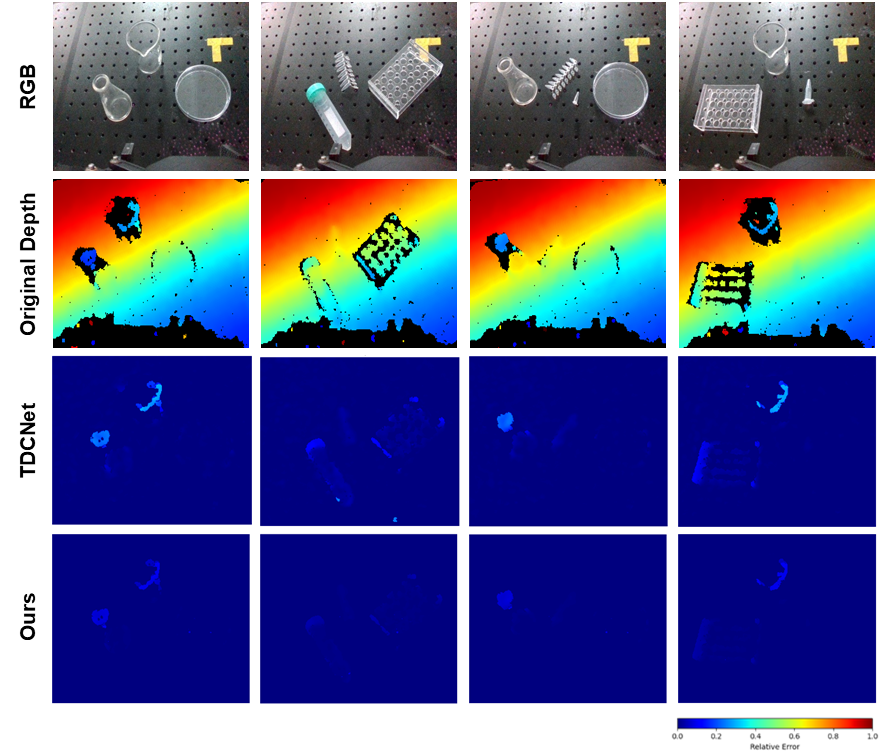} 
\caption{ Qualitative comparison of depth completion. Error heatmaps show that our method maintains lower error distributions and preserves surface curvature better than the baseline TDCNet.}
\label{5}
\end{figure}

Figure~\ref{5} visualizes reduced error accumulation and mitigated geometric collapse, providing more reliable normals for force-closure grasping.

\subsubsection{Geometric Quality Evaluation of Grasp Poses}
To isolate pose quality, we test on \textit{opaque objects} and evaluate whether grasps are well-aligned rather than merely successful.

\textbf{Metrics.}
We compare Top-1 grasps to the ideal principal-axis frame using:
(i) \textbf{Angular Error ($\theta_{err}$)} between approach vector and surface normal (smaller is more vertical);
(ii) \textbf{Center Offset ($d_{off}$)} between grasp center and object centroid along height (smaller is more stable).

\begin{table}[!t]
\caption{Geometric Quality Comparison of Top-1 Grasp Poses}
\label{tab:grasp_quality}
\centering
\begin{tabular}{l c c c c}
\hline
Strategy & Score Basis & Succ. (\%) & Angle Err ($^\circ$) & Offset (mm) \\
\hline
Baseline & Visual Conf. & 94.0 & 22.5 $\pm$ 8.4 & 35.2 $\pm$ 5.1 \\
Ours & Geo. + Phy. & \textbf{98.0}$^{*}$ & \textbf{3.8 $\pm$ 1.5}$^{***}$ & \textbf{8.5 $\pm$ 1.2}$^{***}$ \\
\hline
\multicolumn{5}{l}{\scriptsize * $p<0.05$, *** $p<0.001$. Standard deviations are denoted by $\pm$.}
\end{tabular}
\end{table}

As shown in Table~\ref{tab:grasp_quality}, the baseline often prefers tilted edge grasps (high visual confidence) leading to large angle/offset errors, which is undesirable for precise placement. Our scoring explicitly enforces near-orthogonal, centroid-aligned grasps, achieving 3.8$^\circ$ angular error and 8.5\,mm offset.

\begin{figure}[!t]
\centering
\includegraphics[width=3.5in]{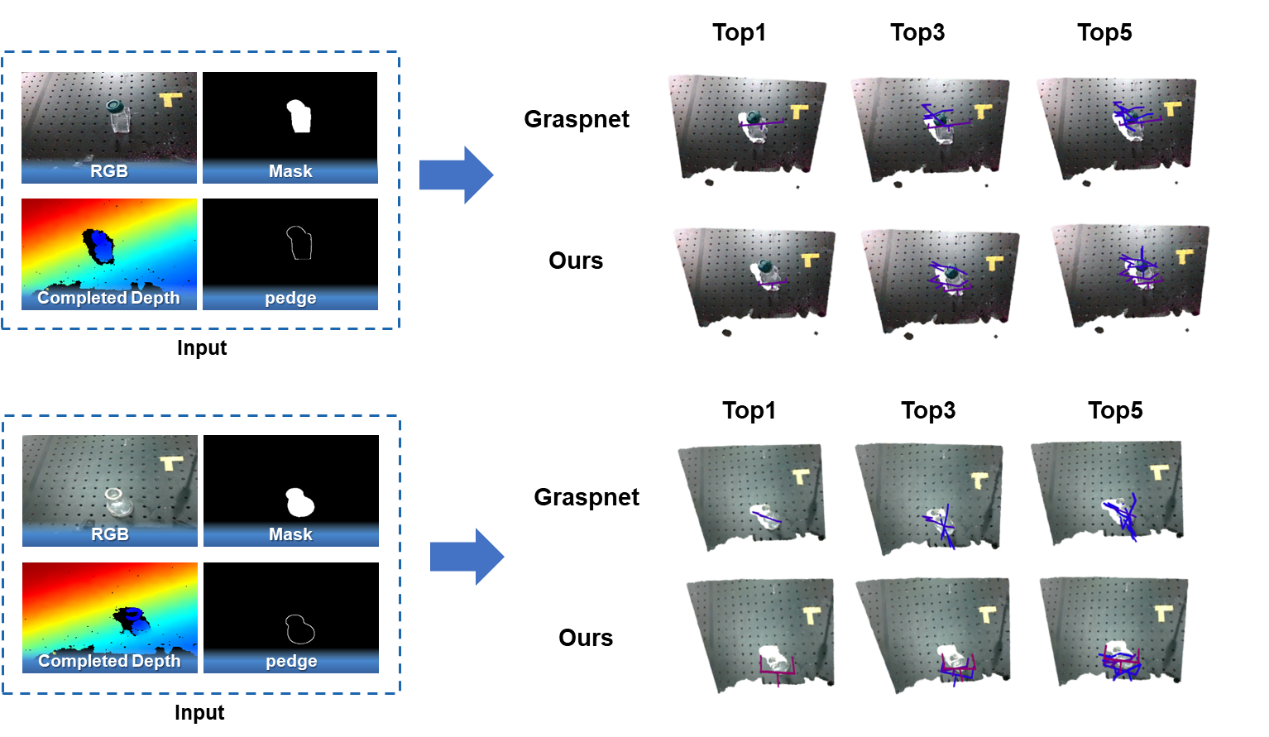} 
\caption{Visualization of Top-$k$ grasp ranking. Our geometry-physics aware re-ranking prioritizes structurally aligned and stable poses over the tilted or edge-biased grasps favored by GraspNet.}
\label{6}
\end{figure}

Figure~\ref{6} further shows that our method ranks structurally aligned grasps higher, functioning as an \textbf{optimizer} rather than a simple failure filter.

\subsection{Comparison with State-of-the-Art Methods}
We evaluate generalization on ClearGrasp under comparable backbones/architectures.

\subsubsection{Transparent Object Segmentation}
We compare with TransLab \cite{xie2020segmenting} and PointRend (ResNet backbones).

\begin{table}[!t]
\caption{Comparison of Instance Segmentation Methods on ClearGrasp}
\label{tab:sota_seg}
\centering
\begin{tabular}{l c c c c}
\hline
Method  & $AP_{mask}$ & $AP_{box}$ & Boundary F \\
\hline
Mask R-CNN  & 76.9 & 78.4 & 52.1 \\
PointRend  & 78.5 & 80.1 & 61.4 \\
TransLab  & \textbf{79.8} & \textbf{81.2} & 58.6 \\
Ours  & 78.2 & 79.5 & \textbf{65.1} \\
\hline
\end{tabular}
\end{table}

Table~\ref{tab:sota_seg} shows that while TransLab slightly improves global AP, our method achieves the best \textbf{boundary quality} (65.1\%), which is more crucial for downstream geometric constraints in grasping.

\subsubsection{Depth Completion Performance}
We compare with ClearGrasp \cite{sajjan2020clear} and NLSPN on ClearGrasp Test-Real.

\begin{table}[!t]
\caption{Comparison of Depth Completion Methods on ClearGrasp}
\label{tab:sota_depth}
\centering
\begin{tabular}{l c c c}
\hline
Method & RMSE (m) $\downarrow$ & MAE (m) $\downarrow$ & $\delta < 1.25$ $\uparrow$ \\
\hline
ClearGrasp & 0.054 & 0.043 & 85.3\% \\
NLSPN & 0.048 & 0.038 & 89.6\% \\
TDCNet (Base) & 0.045 & 0.037 & 90.2\% \\
Ours & \textbf{0.043} & \textbf{0.035} & \textbf{91.5\%} \\
\hline
\end{tabular}
\end{table}

As shown in Table~\ref{tab:sota_depth}, our method achieves the best RMSE and the highest $\delta < 1.25$ (91.5\%), indicating stronger robustness against outliers and reducing collision risks in manipulation.

\subsection{Real-World Robotic Grasping and Manipulation Experiments}
We validate TransGraspNet on a real ``Robot Scientist'' platform to measure end-to-end reliability on transparent glassware.

\subsubsection{Experimental Setup and Protocol}
\textbf{Hardware:} AUBO i5 6-DoF arm with CTAG2F90C adaptive parallel gripper and an eye-in-hand Intel RealSense D435i RGB-D camera.

\textbf{Success Criteria:} A trial is successful only if it completes the full pipeline:
\begin{enumerate}
    \item \textbf{Perception success:} correct identification and localization.
    \item \textbf{Grasp success:} lift to 30\,cm and hold for 3\,s without slipping.
    \item \textbf{Transport success:} horizontal move of 20\,cm to placement area.
    \item \textbf{Placement success:} upright, stable placement without tipping/shaking.
\end{enumerate}

\subsubsection{Quantitative Evaluation of Grasp Stability}
We perform 50 trials per scenario (100 total). Table~\ref{tab:real_world} reports success rates.

\begin{table}[!t]
\caption{Success Rates in Different Scenarios}
\label{tab:real_world}
\centering
\begin{tabular}{l l c c c}
\hline
Scenario & Complexity & Attempts & Success & Rate (\%) \\
\hline
Scenario-I & Simple & 50 & 48 & 96.0 \\
Scenario-II & Hard & 50 & 43 & 86.0 \\
\hline
Average & - & 100 & 91 & 91.0 \\
\hline
\end{tabular}
\end{table}

As shown in Table~\ref{tab:real_world}, TransGraspNet achieves 96.0\% in simple scenes and 86.0\% in clutter. Since success requires \textit{stable placement}, the results also verify that the physics-aware scoring rejects grasps that are pickable but not placeable, improving closed-loop reliability.

\begin{figure}[!t]
\centering
\includegraphics[width=3.0in]{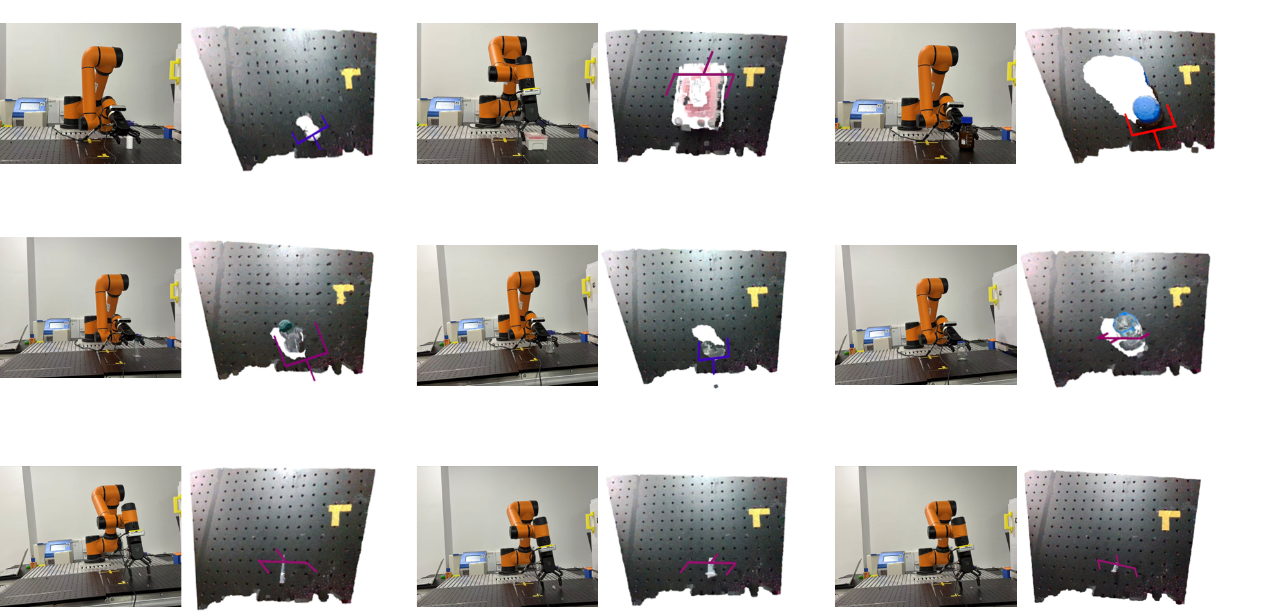} 
\caption{Real robot grasping in single-object scenarios. TransGraspNet reliably predicts and executes high-quality grasps across varying optical properties and scales for both transparent and non-transparent objects.   }
\label{fig:depth_vis}
\end{figure}

\begin{figure}[!t]
\centering
\includegraphics[width=3.0in]{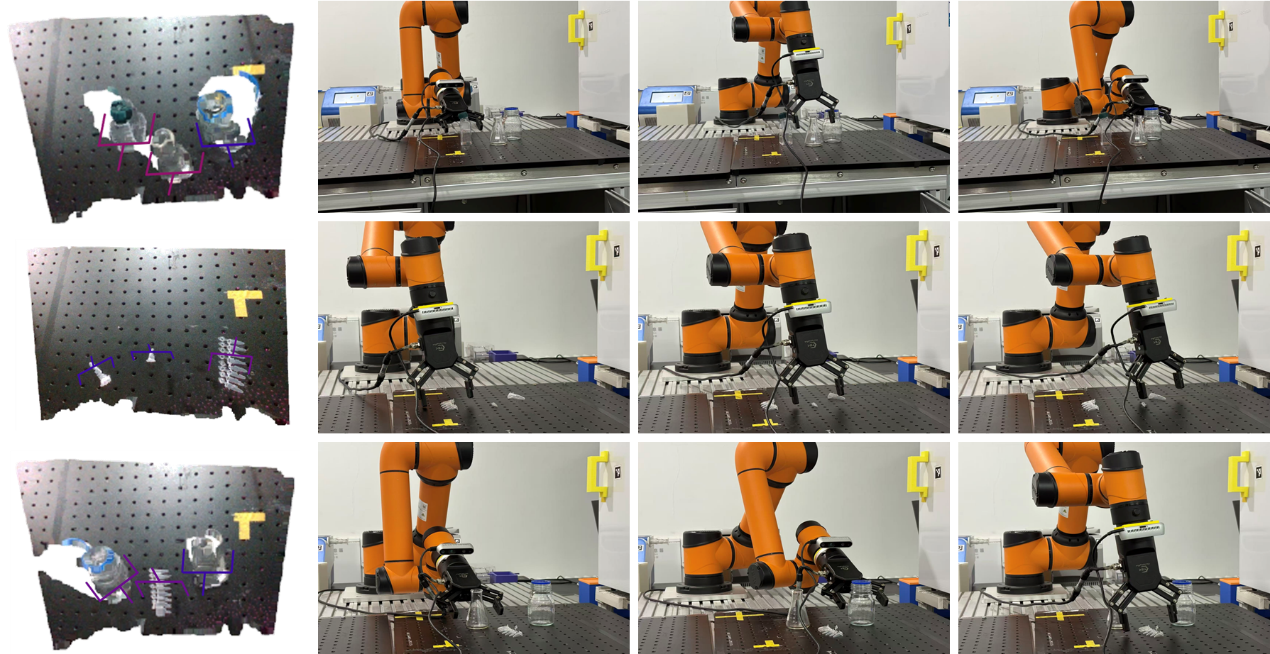} 
\caption{Manipulation results in multi-object scenarios. The complete pick-transport-place sequence demonstrates stable grasping and collision-free execution under clutter and occlusions. 
}
\label{fig}
\end{figure}

\subsubsection{Dynamic Stability Test: Liquid Transport}
We further test dynamic stability via high-speed transport of partially filled glassware, where small pose errors can be amplified into sloshing/spillage.

\begin{figure}[t]
    \centering
     \includegraphics[width=\columnwidth]{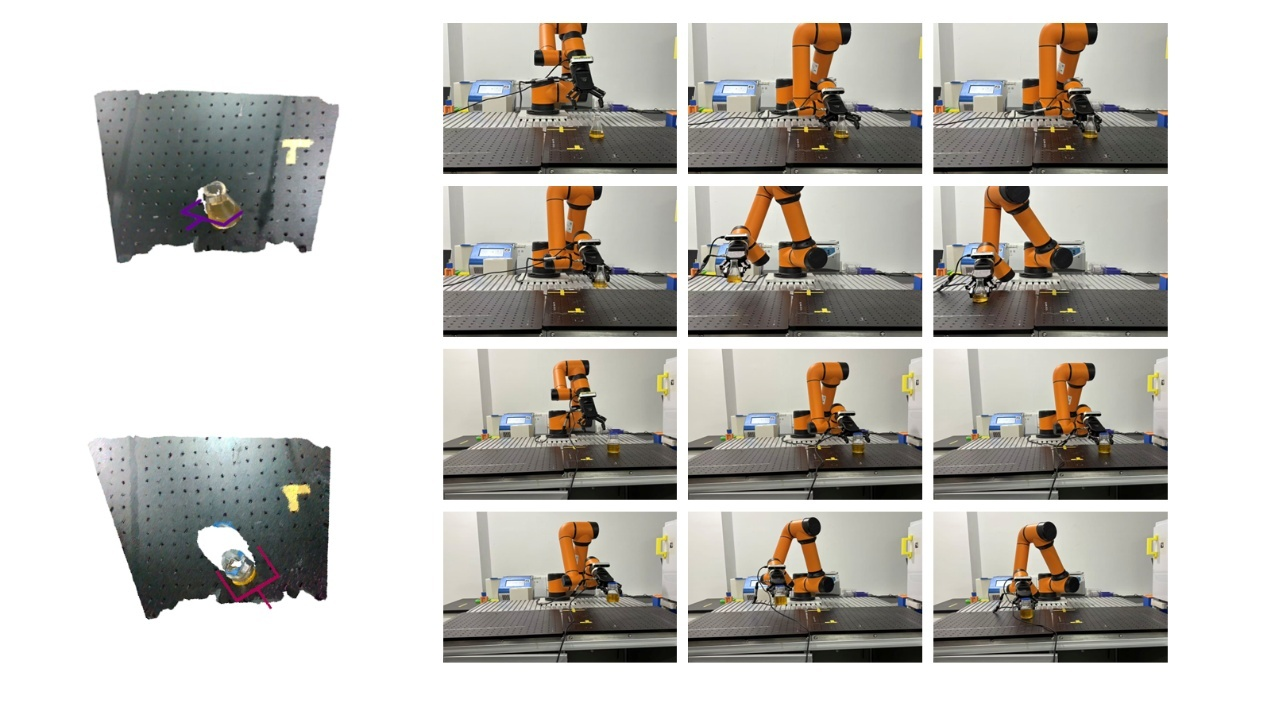}
\caption{Dynamic stability test in high-speed liquid transport. The robot achieves zero spillage during $0.5\text{ m/s}$ transport, confirming that wrench-constrained scoring ensures orientation alignment and suppresses slippage.}
    \label{fig:liquid_transport}
\end{figure}

We use a 50\,ml Erlenmeyer flask and a 100\,ml glass bottle, each filled to 50\% with colored liquid. The robot executes a horizontal transport trajectory of 30\,cm with $v = 0.5\,\text{m/s}$ and $a = 1.0\,\text{m/s}^2$. As shown in Figure~\ref{fig:liquid_transport}, the grasp orientation remains aligned with gravity during lifting and transport; the liquid surface shows only minor fluctuations, and no slippage is observed. The task achieves \textit{zero spillage} throughout.

This improvement is mainly due to wrench-constrained scoring that favors contact points near the centroid, reducing inertial moments, and frictional closure constraints that enlarge disturbance coverage, suppressing micro-slippage on smooth glass surfaces. Compared with purely geometric grasp selection, our approach is more robust to start-stop inertial disturbances, meeting safety requirements in chemical operations.

\section{Conclusion and Future Work}

Targeting the challenges of autonomous robotic manipulation of transparent glassware in unstructured laboratory environments, this paper proposes and validates a full-stack perception and grasping system named \textbf{TransGraspNet}. This research is dedicated to systematically resolving three core challenges: the \textit{Visual Feature Sparsity and Background Coupling} of transparent objects under uncontrolled lighting, the \textit{Depth Incompleteness and Distortion} in 3D geometric reconstruction, and the \textit{Contact Instability} during physical interaction.

To overcome these hurdles, we first proposed an edge-guided perception mechanism. By introducing the E-CBAM attention module and a dual-stream edge branch, we explicitly utilize high-frequency contour features as strong supervision signals, effectively decoupling object boundaries from strong reflections and background clutter, thus solving the ``visual invisibility'' segmentation problem. Building on this, we designed the \textbf{TransGraspNet-Depth} completion network. Utilizing the Edge-Guided Attention Gate (EGAG) and geometric retention loss, it fundamentally repairs depth voids and geometric collapse caused by refraction and transmission, providing the robot with a high-fidelity 3D operational space. Crucially, we innovatively constructed a joint ``Geometric-Physics'' scoring strategy. Transcending traditional visual confidence assessment, it enforces stable grasp poses compliant with friction closure principles through principal axis alignment constraints and Wrench Space analysis.

Extensive experiments on the RobotSci-Glass dataset and a real physical platform demonstrate the system's effectiveness. Results show that even in challenging multi-object occlusion and strong reflection scenarios, the system maintains a high grasp success rate of 86.0\% and achieves zero-spillage dynamic stability in high-speed liquid transport tasks. This verifies that introducing geometric consistency restoration and physical constraints into the planning layer significantly enhances robustness when handling transparent, fragile, and high-risk chemical vessels, laying a solid foundation for automated laboratory systems with ``Robot Scientist'' capabilities.

\printbibliography

\end{document}